\documentclass[10pt,twocolumn,letterpaper]{article}

\usepackage{cvpr}      % To produce the REVIEW version (anonymous, line numbers, page numbers)
\usepackage{graphicx}
\usepackage{amsmath}
\usepackage{amssymb}
\usepackage{booktabs}
\usepackage{multirow}
\graphicspath{{figures/}}

\definecolor{cvprblue}{rgb}{0.21,0.49,0.74}
\usepackage[pagebackref,breaklinks,colorlinks,allcolors=cvprblue]{hyperref}

\def\confYear{2027}
\title{TamperCert: A Benchmark for Detecting Text Tampering in Degree Certificates}

\author{Chi Fung Kwan$^{1}$ \and Kangzhong Wang$^{1}$ \and Youqian Zhang$^{1}$ \and Weijie Chen$^{1}$ \and Jiaqi Gao$^{2}$ \and Yongjian Wu$^{3}$ \and Eugene Yujun Fu$^{2}$ \and Qingqing Ye$^{1}$ \and Haibo Hu$^{1}$\\
$^{1}$ The Hong Kong Polytechnic University\\ 
$^{2}$ The Education University of Hong Kong\\ 
$^{3}$ Independent Researcher\\ 
{\texttt cfukwan@polyu.edu.hk}
}

\begin{document}
\maketitle
\begin{abstract}
Degree certificates play a critical role in high-stakes decisions such as university admission, immigration, and hiring.
The issuing institution vouches for the holder, so a forged certificate may let its bearer pass these checks.
Degree certificate forgery is increasingly prevalent. 
Yet, many assume that it is easy to spot.
Reported cases show otherwise: in 2024, the business school of the University of Hong Kong found that about 30 master's students had enrolled with forged academic credentials.
Few studies have looked at how to detect forged degree certificates, partly because public benchmarks are scarce.
We present TamperCert, a benchmark built from 400 real degree certificates issued by 38 universities.
Each certificate was subjected to 11 publicly available tampering methods to alter the holder's name and further processed with four additional post-processing techniques, resulting in over 20,000 forged images.
We evaluate 13 detectors, a commercial multimodal large language model (MLLM), and detectors trained on TamperCert.
Our findings reveal that no detector is reliable for every tampering method. 
Consequently, a determined forger may still evade detection. 
Moreover, the commercial MLLM misses most forgeries made by general-purpose generative models, and post-processing, such as resizing an image or photographing it with a camera, severely degrades detection.

\end{abstract}
\section{Introduction}
\label{sec:intro}

Degree certificates directly affect admission to further study, professional qualification, visa applications, and employment.
An application usually starts with an electronic copy or a photograph of the certificate, and admissions offices, employers, and immigration authorities have to decide whether it is genuine.
Forged documents are common in these procedures.
In Canada, more than 650,000 letters of acceptance of international students were checked with the issuing institutions in 2024, and about 14,000 of them did not match the records of the institution~\cite{canada-CIMM}.
In July 2024, the business school of the University of Hong Kong reported that about 30 master's students had enrolled with forged academic credentials~\cite{SCMP-HKU}.

Although a forged degree certificate may not be tied to a financial transaction in the way a forged check or contract is, it can lead to unfair admission, hiring, and immigration decisions.
Blockchain-based verification has been proposed~\cite{cheng2018blockchain,pu2023benefits,al2024blockchain}, but many institutions have not adopted it, and certificates issued before such systems existed are not recorded in them.
Manual inspection is unreliable, since most reviewers are not forensic experts.
Contacting the issuing institution is possible, but it is tedious and hard to scale, and a forger can supply fabricated contact details, such as a phone number, an email address or a verification website, that lead the verifier to an accomplice.
Detecting forgery from the image of the certificate therefore remains necessary.

%Unlike checks or contracts, forging degree certificates may not lead to a subsequent financial trail, so it is possibly why there are few studies on certificate . 
%The human examiner can only determine the authenticity based on the static file itself. One critical issue is that the human examiner in institutions may not be professional forensic expert, and it is not guaranteed that they can spot the forgery While digital evidence technologies like blockchain have become mainstream solutions in recent years, they cover only certificates issued by a limited number of institutions and lack a unified verification interface across universities and issuing bodies globally. Manual checking with the original school (that issued the certificate) by different means such as through email, fax, or phone call is feasible, but it is often limited and tedious, making it difficult to complete immediately and one by one during peak application periods. Under time pressure, some employers may simplify or even abandon verification. Visual evidence collection methods capable of rapid and automated screening of certificate images are in practical demand. 
%It is important to emphasize that image detection cannot replace official verification by the issuing institution, but it can provide supplementary judgment before verification is completed.

\begin{figure}[t]
\centering
\includegraphics[width=\linewidth]{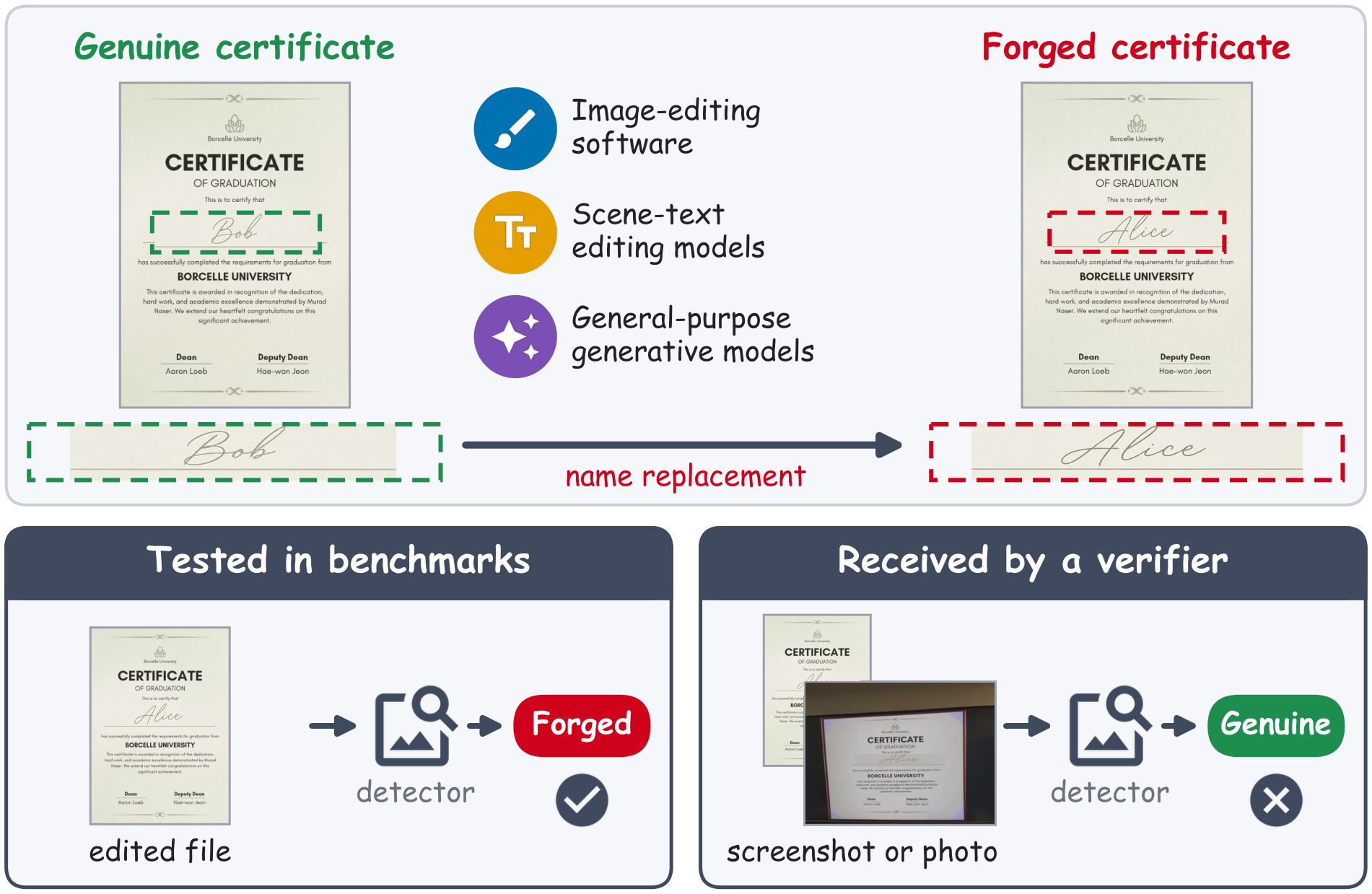}
\caption{Text tampering on a degree certificate, illustrated with a sample certificate. The holder's name is replaced, and a verifier may receive a screenshot or a phone photograph, not the edited file on which detectors are usually tested.}
\label{fig:concept}
\end{figure}

Tampering with the text of a genuine certificate is enough to forge one, because the result keeps its seals, signatures and layout.
Such tampering was long done by hand in image-editing software, and recent editing models require far less expertise~\cite{qu2026dancetext}.
Scene-text editing models rewrite a word in its original font and background~\cite{wu2019srnet,ji2024diffste,zhao2024udifftext,chen2024textdiffuser2,tuo2024anytext2}, and general-purpose generative models edit the text of a document from a one-line instruction, inside a selected region or by regenerating the full image~\cite{wu2025qwenimage}.
It is essential to note that, before a forged certificate is submitted, it may also be saved again, captured in a screenshot or photographed, so the verifier may not see the edited file (as shown in the bottom-right part of \cref{fig:concept}).

Image and document forensics have matured in recent years, but existing benchmarks do not directly cover degree certificates.
The dataset~\cite{tornes2023receipt}, for example, provides 988 scanned receipts, 163 of which are forged, and shows that automated detection is feasible for a specific type of document, but receipts differ from certificates in their layout and in the text that is tampered.
Other tampered-text benchmarks mainly target scene text and documents such as contracts and invoices~\cite{wang2022tic13,qu2023doctamper,luo2024rtm,qu2025ostf,yu2025fsts}, and recent ones add general-purpose generative models for many types of text images~\cite{qu2026textshield,qu2026dancetext}.
Certificates are rarely included~\cite{zhuang2023reloc}, and detectors are normally tested on edited files, with robustness examined mainly through JPEG recompression and synthetic distortions~\cite{qu2023doctamper,wong2025adcdnet}.
Detection in screenshots and phone photographs has received little attention.

To bridge this gap, we introduce TamperCert, a benchmark for detecting text tampering in degree certificates.
It starts from 400 genuine certificates issued by 38 universities and changes only the holder's name.
On every certificate, 11 tampering methods write the same target name.
They cover image-editing software, scene-text editing models and general-purpose generative models, and either edit only the name region (regional edits) or regenerate the full image (full-image regeneration).
Every genuine and forged image is then provided as a lossless file, a JPEG file, a resized JPEG file, a screenshot and a phone photograph, which gives 21,990 forged images, and both classes are processed in the same way.

We evaluate 13 detectors from three families, namely tampering localizers, AI-generated image detectors and a multimodal large language model (MLLM), together with a commercial MLLM and detectors trained on TamperCert.
The aim is to represent the main families of methods, not to cover every state-of-the-art system.
The results show that TamperCert remains challenging: on JPEG files at least one detector reaches an AUC of 0.90 for eight of the 11 tampering methods, but in phone photographs for only two.
Our main contributions are as follows.
\begin{itemize}
    \item We introduce TamperCert, a benchmark for detecting text tampering in degree certificates. To our knowledge, it is the first tampered-text benchmark in which image-editing software, scene-text editing models and general-purpose generative models make the same edit on the same document.
    \item We benchmark 13 detectors, a commercial MLLM and detectors trained on TamperCert. No detector is reliable for both regional edits and full-image regeneration, and the commercial MLLM misses most forgeries made by general-purpose generative models.
    \item We observe that detection depends on how the image has been processed. After resizing alone, tampering localizers miss most regional edits, and in phone photographs names replaced with Photoshop or general-purpose generative models are not reliably detected by any detector.
\end{itemize}

\section{Related Work}
\label{sec:related}

\begin{table*}[t]
\centering
\small
\setlength{\tabcolsep}{4pt}
\caption{Comparison with tampered-text datasets. ``Software'': manual or rule-based editing; ``Scene-text'': scene-text editing models; ``Generative'': general-purpose generative models; ``Full image'': the entire image is regenerated; ``Screenshot, photo'': forgeries also provided as screenshots or photographs. $^{\dagger}$Commercial apps only.}
\label{tab:dataset_comparison}
\resizebox{\textwidth}{!}{%
\begin{tabular}{@{}lllccccc@{}}
\toprule
 & & & \multicolumn{3}{c}{Tampering methods} & & \\
\cmidrule(lr){4-6}
Dataset & Images & Edited text & Software & Scene-text & Generative & Full image & Screenshot, photo \\
\midrule
Tampered-IC13~\cite{wang2022tic13} & scene text & text instances & -- & \checkmark & -- & -- & -- \\
DocTamper~\cite{qu2023doctamper} & contracts, invoices, receipts & text selected by OCR & \checkmark & -- & -- & -- & -- \\
RTM~\cite{luo2024rtm} & documents, text images & text instances & \checkmark & -- & -- & -- & -- \\
OSTF~\cite{qu2025ostf} & scene text & text instances & -- & \checkmark & -- & -- & -- \\
FSTS~\cite{yu2025fsts} & text images & text instances & \checkmark & -- & -- & -- & -- \\
TFR~\cite{qu2026textshield} & documents, identity cards, scene text & text regions & \checkmark & \checkmark & \checkmark & \checkmark & -- \\
DanceText~\cite{qu2026dancetext} & 108 types of text images & chosen by a multimodal model & \checkmark$^{\dagger}$ & \checkmark & \checkmark & \checkmark & -- \\
\midrule
TamperCert (ours) & real degree certificates & holder's name, same target & \checkmark & \checkmark & \checkmark & \checkmark & \checkmark \\
\bottomrule
\end{tabular}}
\end{table*}

\paragraph{Tampered-text datasets.}
Early datasets were built with a single category of tampering method.
Tampered-IC13~\cite{wang2022tic13} edits scene text with a scene-text editing model, DocTamper~\cite{qu2023doctamper} synthesizes tampered documents by copy-move, splicing and text rendering, and RTM~\cite{luo2024rtm} and FSTS~\cite{yu2025fsts} are made with Photoshop, operated by hand or by script.
OSTF~\cite{qu2025ostf} applies eight scene-text editing models to scene text and shows that detectors trained on some of them do not transfer to the others.
Two recent datasets add general-purpose generative models.
TFR~\cite{qu2026textshield} combines traditional methods, scene-text editing models and fully generated images, and DanceText~\cite{qu2026dancetext} covers 108 types of text images forged by 45 generators, from scene-text editing models to commercial multimodal models and commercial apps.
Datasets of identity documents and other credentials use mock or synthetic documents, since real ones carry personal data and are protected by legal and security restrictions~\cite{bulatov2022midv2020,boned2024sidtd,korshunov2025fantasyid,li2026cognicred}.
Certificates have appeared in Certificate PS, a set of certificate images that were captured by phone and then tampered by hand~\cite{zhuang2023reloc}.
TamperCert consists of real degree certificates, applies methods of all three categories to the same text with the same target name, and provides each forgery and its genuine certificate also as a screenshot and a phone photograph (\cref{tab:dataset_comparison}).

\paragraph{Forgery detectors.}
Tampering localizers predict which pixels of an image were modified, from a learned noise fingerprint~\cite{guillaro2023trufor}, from DCT coefficients~\cite{kwon2022catnet,qu2023doctamper,wong2025adcdnet}, or from the distance of text features to a learned representation of genuine text~\cite{qu2025ostf}.
AI-generated image detectors classify whole images.
Some rely on CLIP features~\cite{ojha2023univfd} or up-sampling traces~\cite{tan2024npr}, and others are trained on images from thousands of generators~\cite{park2025commfor} or on real images that a generator has regenerated or reconstructed, so that the two classes match in content and format~\cite{guillaro2025bfree,rajan2025aligned,chen2025dda}.
MLLMs have been fine-tuned to explain and localize forgeries~\cite{xu2025fakeshield,huang2025sida,qu2026textshield} and prompted directly as zero-shot detectors~\cite{jia2024chatgpt}.
For natural images it is known that localizers detect inpainted regions but not images that are fully regenerated~\cite{mareen2024tgif}, and that AI-generated image detectors can exploit format and size differences between the classes~\cite{grommelt2024fakeorjpeg,guillaro2025bfree,rajan2025aligned}.
We test whether these findings hold for the same edit on the same document, and in screenshots and phone photographs.

\paragraph{Robustness to post-processing.}
Document forensics has examined robustness mostly through JPEG recompression and synthetic distortions such as noise, blur and downscaling~\cite{qu2023doctamper,wong2025adcdnet}.
Recapture is usually treated as a separate task, for instance deciding whether a document image was photographed from a screen~\cite{chen2024cma}, and tampering localization after printing and scanning has also been studied~\cite{chen2025unmask}.
We instead test whether a forged certificate can be told apart from a genuine one when both are screenshots or phone photographs.

\section{The TamperCert Benchmark}
\label{sec:benchmark}

\begin{figure*}[t]
\centering
\includegraphics[width=\textwidth]{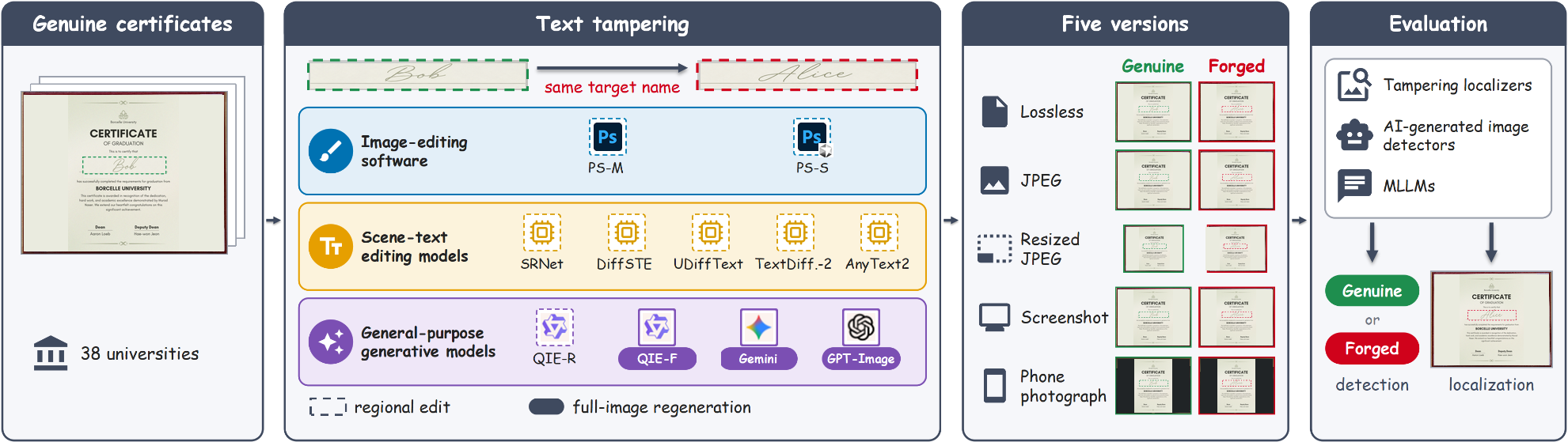}
\caption{Overview of TamperCert, illustrated with a sample certificate: genuine certificates, the 11 tampering methods that write the same target name, the five versions of every image, and the evaluation.}
\label{fig:overview}
\end{figure*}

The task in TamperCert is to decide, from a single image of a degree certificate, whether its text has been tampered with and, if so, where (\cref{fig:overview}).
On each of the 400 genuine certificates, 11 tampering methods replace the holder's name, and every genuine and forged image is provided in five versions.

\subsection{Genuine Certificates}
\label{sec:collection}
The benchmark starts from 400 degree certificates issued by 38 universities.The source images were collected from publicly available, unaltered certificates, following the order of the QS rankings. 
We call them ``genuine certificates'' in this work.
%the collection includes versions with and without QR codes. Data acquisition ceased once 400 images had been gathered, covering a total of 38 institutions.
The images have a median size of about one megapixel.
Certificates of the same university share a layout, and the number of certificates per university ranges from 1 to 34.
For every certificate, we annotate the holder's name and its bounding box, which we call the name region, and verify all annotations manually.
The name is a small target, and its region covers a median of 1\% of the page area.

\subsection{Tampering Methods}
\label{sec:editing}
For each certificate, we fix one target name by changing one word of the holder's name, in most cases by one to three characters, so that the target name stays plausible and keeps the format of the original.
All methods write this same target name.
The 11 methods belong to three categories, each of which can replace a name without training a model (\cref{tab:methods}).

\paragraph{Image-editing software.}
In the manual edit, an annotator removes the name in Photoshop, restores the background and types the target name in the closest matching font, size and color.
The scripted edit automates the same steps from the name region.

\paragraph{Scene-text editing models.}
Five models receive a crop around the name together with the target name and return the crop with the text replaced, which is pasted back into the page.
They include the GAN-based SRNet~\cite{wu2019srnet} and four diffusion-based models, DiffSTE~\cite{ji2024diffste}, UDiffText~\cite{zhao2024udifftext}, TextDiffuser-2~\cite{chen2024textdiffuser2} and AnyText2~\cite{tuo2024anytext2}.

\paragraph{General-purpose generative models.}
These models edit an image from a text instruction, which states the original name and the target name.
We use the open-source Qwen-Image-Edit~\cite{wu2025qwenimage} in two ways, on a crop around the name that is pasted back (QIE-R) and on the full image (QIE-F).
The commercial Gemini~\cite{google_gemini_image} and GPT-Image~\cite{openai_gpt_image} are applied to the full image.

\begin{figure}[t]
\centering
\includegraphics[width=\linewidth]{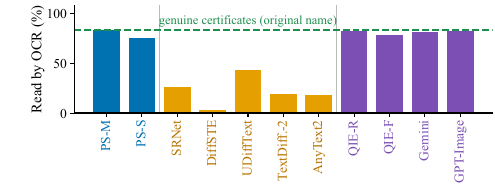}
\caption{Share of forgeries in which OCR reads the target name correctly. The green dashed line is the share for the original names of genuine certificates.}
\label{fig:dataset_stats}
\end{figure}

After generation, we manually check the outputs and exclude two, one from UDiffText and one from AnyText2, in which the name is not changed.
An OCR engine reads names written by image-editing software and general-purpose generative models almost as often as genuine names, whereas the scene-text editing models often produce visible flaws such as distorted characters (\cref{fig:dataset_stats}).

\begin{table}[t]
\centering
\small
\setlength{\tabcolsep}{2.2pt}
\caption{The 11 tampering methods in TamperCert. Regional-editing methods change only a crop around the name, and full-image methods regenerate the entire certificate.}
\label{tab:methods}
\begin{tabular}{@{}llll@{}}
\toprule
Category & Method & Abbrev. & Scope \\
\midrule
\multirow{2}{*}{\shortstack[l]{Image-editing\\software}} & Photoshop, manual & PS-M & regional \\
 & Photoshop, scripted & PS-S & regional \\
\midrule
\multirow{5}{*}{\shortstack[l]{Scene-text\\editing models}} & SRNet~\cite{wu2019srnet} & & regional \\
 & DiffSTE~\cite{ji2024diffste} & & regional \\
 & UDiffText~\cite{zhao2024udifftext} & & regional \\
 & TextDiffuser-2~\cite{chen2024textdiffuser2} & TextDiff.-2 & regional \\
 & AnyText2~\cite{tuo2024anytext2} & & regional \\
\midrule
\multirow{4}{*}{\shortstack[l]{General-purpose\\generative\\models}} & Qwen-Image-Edit~\cite{wu2025qwenimage} & QIE-R & regional \\
 & Qwen-Image-Edit~\cite{wu2025qwenimage} & QIE-F & full image \\
 & Gemini~\cite{google_gemini_image} & & full image \\
 & GPT-Image~\cite{openai_gpt_image} & & full image \\
\bottomrule
\end{tabular}
\end{table}

\subsection{Image Versions}
\label{sec:versions}
Every genuine and forged image is provided in five versions, which correspond to the forms in which a certificate image can reach a verifier: the edited file, a file that has been saved again or resized, a screenshot, and a photograph.
The \emph{lossless} version stores the image as PNG, the \emph{JPEG} version compresses it at quality 90, and the \emph{resized JPEG} version scales each side to 75\% before the same compression.
For the \emph{screenshot} version the image is displayed in a web browser at full-HD width (1,920 pixels), which enlarges most images, and captured as a screenshot, and for the \emph{phone photograph} version the image displayed on a monitor is recaptured with a smartphone.
The compression and scaling settings are mild, so the versions do not represent heavy degradation.
The genuine and forged images of a version are produced with the same procedure and settings, and a detector is evaluated within one version, so it cannot separate the classes by file format or compression setting.
Two differences between the classes come from the tampering methods themselves.
A fully regenerated image does not have the compression history of its source, and Gemini and GPT-Image change the size of the image (the supplementary material gives a control with resized genuine certificates).

\section{Experimental Setup}
\label{sec:protocol}

\paragraph{Detectors.}
We evaluate 13 publicly available detectors as released, without training them on certificates.
The six tampering localizers are TruFor~\cite{guillaro2023trufor}, CAT-Net~\cite{kwon2022catnet}, DTD~\cite{qu2023doctamper}, ADCD-Net~\cite{wong2025adcdnet}, DAF~\cite{qu2025ostf} and error level analysis (ELA)~\cite{krawetz2007ela}.
The six AI-generated image detectors are UnivFD~\cite{ojha2023univfd}, NPR~\cite{tan2024npr}, Community Forensics~\cite{park2025commfor}, B-Free~\cite{guillaro2025bfree}, AlignedForensics~\cite{rajan2025aligned} and DDA~\cite{chen2025dda}.
The MLLM is the open-source Qwen3-VL~\cite{bai2025qwen3vl}, which is asked whether any text in the image has been altered, and the probability of a positive answer is its score.
DTD requires JPEG input and is not evaluated on the lossless version.
In addition, we query a commercial MLLM, Claude (Opus 5.5, accessed in November 2026), for which we report the share of images that it flags, and we train a localizer and an image-level classifier on part of TamperCert (\cref{sec:finetune}).

% Generated by figs_src/make_tables.py from EXPERIMENT_DATA_1111.md (Secs. 2.1 and 8.2). Do not edit numbers by hand.
\begin{table*}[t]
\centering
\footnotesize
\setlength{\tabcolsep}{4.6pt}
\caption{Image-level AUC of the 13 detectors on JPEG files (top) and phone photographs (bottom), against the genuine certificates of the same version. ``Min'' is the lowest AUC over the 11 methods, and the best result per method is in bold. The 95\% confidence intervals have a median half-width of 0.03 (see the supplementary material).}
\label{tab:main_results}
\begin{tabular}{@{}l cc ccccc c ccc c@{}}
\toprule
 & \multicolumn{2}{c}{\shortstack{Image-editing\\software}} & \multicolumn{5}{c}{Scene-text editing models} & \multicolumn{4}{c}{General-purpose generative models} & \\
\cmidrule(lr){2-3}\cmidrule(lr){4-8}\cmidrule(lr){9-12}
 & \multicolumn{8}{c}{regional edits} & \multicolumn{3}{c}{full-image regeneration} & \\
\cmidrule(lr){2-9}\cmidrule(lr){10-12}
Detector & PS-M & PS-S & SRNet & DiffSTE & UDiffText & TextDiff.-2 & AnyText2 & QIE-R & QIE-F & Gemini & GPT-Image & Min \\
\midrule
\multicolumn{13}{@{}l}{\emph{JPEG file}} \\
TruFor & .71 & .76 & .83 & .82 & .72 & .80 & .80 & .71 & .34 & .41 & .23 & .23 \\
CAT-Net & .81 & .81 & .90 & .88 & .83 & .89 & .91 & .76 & .60 & .63 & .49 & .49 \\
DTD & \textbf{.92} & \textbf{.94} & .96 & \textbf{.98} & \textbf{.96} & \textbf{.97} & \textbf{.98} & \textbf{.94} & .70 & \textbf{.82} & \textbf{.74} & .70 \\
ADCD-Net & .76 & .85 & .68 & .69 & .59 & .61 & .78 & .70 & .56 & .67 & .56 & .56 \\
DAF & .58 & .76 & .90 & .85 & .79 & .72 & .73 & .71 & .67 & .49 & .59 & .49 \\
ELA & .50 & .52 & .51 & .49 & .50 & .51 & .52 & .49 & .48 & .64 & .49 & .48 \\
\midrule
UnivFD & .52 & .54 & .62 & .65 & .66 & .61 & .66 & .52 & .51 & .53 & .55 & .51 \\
NPR & .49 & .49 & .49 & .49 & .50 & .50 & .49 & .50 & .45 & .55 & .63 & .45 \\
Community Forensics & .49 & .50 & .53 & .50 & .50 & .52 & .52 & .50 & .64 & .60 & .54 & .49 \\
B-Free & .53 & .53 & .60 & .69 & .68 & .62 & .70 & .54 & \textbf{.74} & .79 & .72 & .53 \\
AlignedForensics & .52 & .50 & .50 & .51 & .50 & .52 & .53 & .50 & .72 & .63 & .42 & .42 \\
DDA & .52 & .48 & .53 & .55 & .53 & .57 & .55 & .51 & \textbf{.74} & .80 & .42 & .42 \\
\midrule
Qwen3-VL & .54 & .65 & \textbf{.98} & \textbf{.98} & .89 & .89 & .94 & .55 & .61 & .57 & .50 & .50 \\
\midrule
\multicolumn{13}{@{}l}{\emph{Phone photograph}} \\
TruFor & .70 & .48 & .49 & .54 & .51 & .47 & .50 & .53 & .48 & .66 & .46 & .46 \\
CAT-Net & .55 & .49 & .53 & .52 & .50 & .52 & .51 & .50 & .47 & .70 & .61 & .47 \\
DTD & .47 & .43 & .43 & .46 & .44 & .45 & .50 & .48 & .40 & .52 & .48 & .40 \\
ADCD-Net & .53 & .58 & .55 & .60 & .56 & .59 & .70 & .54 & .49 & .57 & .59 & .49 \\
DAF & .53 & .55 & .67 & .68 & .62 & .61 & .59 & \textbf{.57} & .56 & .55 & .56 & .53 \\
ELA & .40 & .55 & .55 & .53 & .54 & .54 & .56 & .51 & .56 & .42 & .63 & .40 \\
\midrule
UnivFD & .43 & .54 & .61 & .63 & .66 & .63 & .64 & .49 & .60 & .50 & .59 & .43 \\
NPR & .59 & .54 & .54 & .54 & .56 & .55 & .55 & .51 & .57 & \textbf{.75} & .69 & .51 \\
Community Forensics & .45 & .52 & .56 & .56 & .53 & .55 & .54 & .53 & .51 & .49 & .41 & .41 \\
B-Free & .61 & .51 & .51 & .54 & .55 & .50 & .54 & .47 & .58 & .74 & \textbf{.72} & .47 \\
AlignedForensics & \textbf{.75} & .57 & .57 & .58 & .60 & .55 & .58 & .56 & \textbf{.70} & .62 & .61 & .55 \\
DDA & .54 & .55 & .57 & .54 & .56 & .58 & .57 & .53 & .58 & .63 & .33 & .33 \\
\midrule
Qwen3-VL & .52 & \textbf{.61} & \textbf{.95} & \textbf{.95} & \textbf{.81} & \textbf{.84} & \textbf{.88} & .53 & .56 & .55 & .47 & .47 \\
\bottomrule
\end{tabular}
\end{table*}

\paragraph{Metrics.}
Detection is measured by the area under the ROC curve (AUC) between the forged images of one tampering method and the genuine images of the same version.
Localization is measured with the pointing game~\cite{zhang2016pointing}, in which a forgery counts as a hit if the strongest response of the detector lies inside the name region.
It needs no threshold and applies to response maps and boxes alike.
For Qwen3-VL, the response is the centre of the box that it returns when asked to mark the altered text.
The supplementary material gives confidence intervals from a bootstrap over universities, detection rates at a fixed false-positive rate and implementation details.

\section{Results and Analysis}
\label{sec:experiments}

\subsection{Detection across Tampering Methods}
\label{sec:across_methods}
\textbf{No detector is reliable for both regional edits and full-image regeneration.}
On JPEG files (\cref{tab:main_results}), most tampering localizers detect regional edits, and the best of them, DTD, reaches an AUC of at least 0.92 on all eight regional-editing methods.
They are weaker on full-image regeneration, where there is no boundary between edited and original pixels.
TruFor gives fully regenerated images lower scores than genuine certificates (0.23--0.41).
AI-generated image detectors reach at most 0.70 on regional edits, where only a small part of the page is synthetic, and at most 0.80 on full-image regeneration.
Qwen3-VL detects the five scene-text editing models and little else.
That localizers miss fully regenerated images is known for natural images~\cite{mareen2024tgif}, and here it holds when every column of \cref{tab:main_results} contains the same certificates with the same target name.
A verifier does not know which method a forger has used, so the relevant number is the lowest AUC over the 11 methods (last column), which is 0.70 or lower for every detector.

\begin{figure*}[t]
\centering
\includegraphics[width=\textwidth]{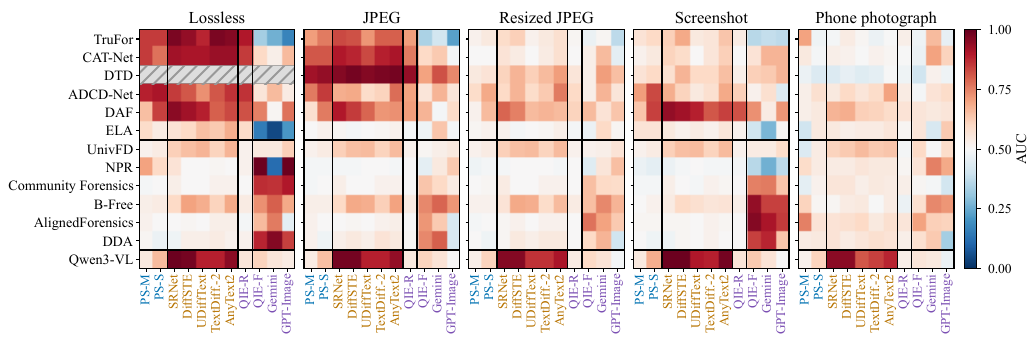}
\caption{Image-level AUC of each detector (rows) on each tampering method (columns) in the five versions. Red: forgeries are detected; blue: they score lower than genuine certificates; hatched: not available.}
\label{fig:auc_versions}
\end{figure*}

\subsection{Detection after Post-processing}
\label{sec:after_processing}
\textbf{Off-the-shelf detectors miss most regional edits after resizing or phone recapture.}
Resizing before compression weakens most localizers on regional edits, and DTD falls from 0.92--0.98 to 0.55--0.71.
In screenshots DAF is higher than on JPEG files, with 0.81 or more on seven of the eight regional-editing methods.
In lossless files AI-generated image detectors reach up to 0.99 on full-image regeneration (\cref{fig:auc_versions}), where only the genuine files keep a compression history (\cref{sec:versions}), and none of them exceeds 0.80 after JPEG compression.
In screenshots B-Free and AlignedForensics improve to 0.84--0.96 on full-image regeneration.
The same increase appears when both classes are simply enlarged by the same factor (see the supplementary material), which suggests that these detectors benefit from the enlargement of the displayed page.
Phone photographs remove this gain (\cref{tab:main_results}).
No detector exceeds 0.75 on text replaced with Photoshop or general-purpose generative models, whether the edit is regional or regenerates the full image.
The number of methods that at least one detector detects with an AUC of 0.90 or more is eight on JPEG files, three on resized JPEG files, six in screenshots and two in phone photographs (\cref{fig:auc_versions}).
The two that remain are scene-text editing models detected by Qwen3-VL.
On the other methods the highest values, 0.70--0.75 for PS-M and the three full-image methods, may reflect differences between the photographs other than the replaced name, since TruFor, for example, reaches 0.70 on PS-M but locates the name in only 0.02 of these photographs.

\begin{figure}[t]
\centering
\includegraphics[width=\linewidth]{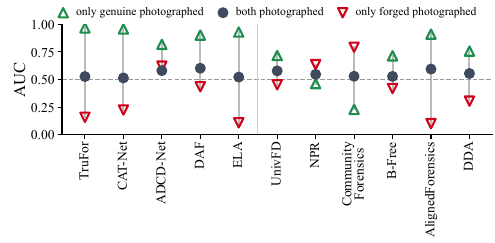}
\caption{Mean AUC of each detector on the eight regional-editing methods when both classes are photographed (circles) and when only one class is photographed and the other is kept as a lossless file (triangles). DTD and Qwen3-VL are not available.}
\label{fig:mismatch}
\end{figure}

\textbf{Post-processing only one class gives misleading results.}
When only the forged images are photographed and the genuine certificates are kept as lossless files, TruFor scores 0.14--0.26 on the regional-editing methods, and when only the genuine certificates are photographed it scores 0.95--0.98, against 0.47--0.70 when both are photographed (\cref{fig:mismatch}).
In the first two settings a detector separates photographs from files, not forged from genuine certificates.

\subsection{Localization of the Tampered Text}
\label{sec:localization}
\begin{figure}[t]
\centering
\includegraphics[width=\linewidth]{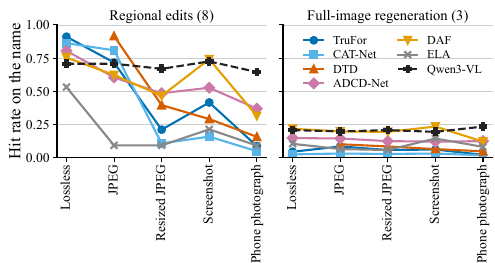}
\caption{Hit rate on the name for the six tampering localizers and Qwen3-VL, averaged over the eight regional-editing methods (left) and the three full-image methods (right).}
\label{fig:localization}
\end{figure}
\textbf{Localization degrades with post-processing and fails for full-image regeneration.}
A verifier also needs to know which part of a certificate was changed (\cref{fig:localization}).
For regional edits, most localizers locate the name in the majority of lossless and JPEG files, but less often in the other three versions (except DAF in screenshots), and none of them exceeds 0.37 in phone photographs.
The hit rate of TruFor, for example, falls from 0.91 on lossless files to 0.09 in phone photographs.
The hit rate of Qwen3-VL, in contrast, stays between 0.65 and 0.73, and its hits come mostly from the scene-text editing models, which suggests that it relies on the appearance of the text.
For full-image regeneration, the average hit rate of every detector is at most 0.23 in every version, which is close to the hit rates on genuine certificates (up to 0.17 on JPEG files), where nothing was changed.
A detector that flags a fully regenerated image therefore gives the verifier no indication of what was changed.

\subsection{Multimodal Large Language Models}
\label{sec:mllm}
\begin{figure}[t]
\centering
\includegraphics[width=\linewidth]{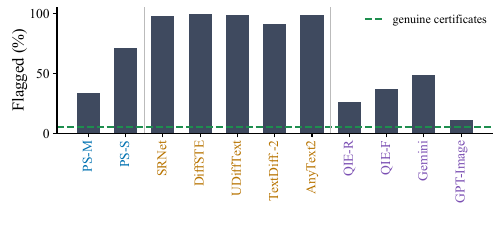}
\caption{Share of forgeries that Claude flags as altered (JPEG version, 94 certificates per method). The green line is the share of genuine certificates that it flags.}
\label{fig:mllm}
\end{figure}
\begin{figure}[t]
\centering
\includegraphics[width=\linewidth]{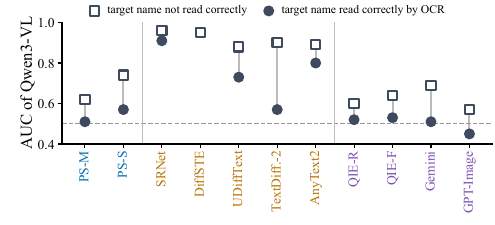}
\caption{AUC of Qwen3-VL in phone photographs, separately for forgeries whose target name OCR reads correctly and for the others. Groups with fewer than 30 forgeries are omitted.}
\label{fig:ocr_split}
\end{figure}
\textbf{The commercial MLLM misses most forgeries of general-purpose generative models.}
We show Claude 94 genuine certificates and their forgeries by the 11 tampering methods in the JPEG version, and ask whether each image has been altered.
It rarely flags genuine certificates (5\%, \cref{fig:mllm}).
It flags at least 91\% of the forgeries made by each scene-text editing model and 71\% of the scripted Photoshop edits, but only 34\% of the manual Photoshop edits and 11--49\% of the forgeries made by general-purpose generative models.
Forgeries made by GPT-Image are flagged in 11\% of the cases, against 5\% of the genuine certificates.
This order follows \cref{fig:dataset_stats}, where the scene-text editing models are the methods that often fail to write the target name.
To test this link within each method, we split its forgeries by whether OCR reads the target name correctly and compute the AUC of Qwen3-VL in phone photographs for both parts (\cref{fig:ocr_split}).
For the five scene-text editing models together, forgeries with a correctly read name are detected with an AUC of 0.76 and the others with 0.92.
A gap also appears on JPEG files (0.86 and 0.96), where the localizer DTD shows no difference (0.98 and 0.97).

\subsection{Training on TamperCert}
\label{sec:finetune}
\begin{figure}[t]
\centering
\includegraphics[width=\linewidth]{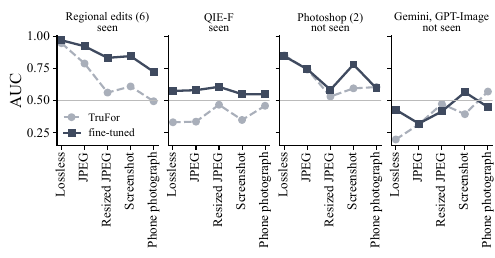}
\caption{AUC of TruFor before (gray, dashed) and after (dark, solid) fine-tuning on TamperCert, on 189 test certificates, for methods that are seen and not seen in training.}
\label{fig:finetune}
\end{figure}
\textbf{Training helps for regional-editing methods seen in training, but not for full-image regeneration.}
To test whether training on certificates closes these gaps, we fine-tune TruFor on forgeries from the five scene-text editing models, QIE-R and QIE-F, and test it on 189 certificates that are not used for training (the split is given in the supplementary material). A single model is trained on crops of the lossless version, with random rescaling and JPEG compression as augmentation, and is tested on all five versions, so it sees no screenshots or phone photographs in training.
For the six regional-editing methods seen in training, detection becomes more robust (\cref{fig:finetune}).
The mean AUC on resized JPEG files rises from 0.56 to 0.83, and it is 0.72 in phone photographs, where the original TruFor is at chance.
The two Photoshop methods, which are not seen, benefit mainly in screenshots.
QIE-F, the full-image method seen in training, reaches only 0.55--0.61, and Gemini and GPT-Image stay at or below 0.59.
A localizer is trained to separate edited from original pixels within one image, so we also train an image-level classifier, a ConvNeXt-T~\cite{liu2022convnext}, on JPEG files of one tampering method at a time.
It detects the full-image method it was trained on in certificates of unseen universities (0.94--1.00), but falls to 0.51--0.61 when it is trained on all other methods instead, and it stays at chance on PS-M and QIE-R even when trained on them (see the supplementary material).

\section{Discussion}
\label{sec:discussion}

\paragraph{Implications.}
Full-image regeneration is the hardest case even before any post-processing: on JPEG files the best detector reaches an AUC of 0.74 on GPT-Image, and Claude flags only 11\% of these forgeries.
After a phone photograph, names replaced with Photoshop or a general-purpose generative model are hard as well, and no detector exceeds 0.75 on them.
What the detectors still find in phone photographs are mainly the forgeries of scene-text editing models, whose text often has visible flaws.
Names written with Photoshop or general-purpose generative models are read by OCR almost as often as genuine names, and comparing a certificate with others of the same university does not expose them either (AUC of at most 0.60 on lossless and JPEG files, see the supplementary material).
For some detectors the size of the name also matters, whereas the number of changed characters matters little.
On JPEG files, the mean AUC of TruFor on regional edits is 0.75 when one character is changed and 0.77 when two or three are changed, but it is 0.67 for the smallest third of the names and 0.86 for the largest third (see the supplementary material).
Benchmarks should therefore report results after genuine and forged images have gone through the same post-processing, because results on edited files overstate what a detector can do for a verifier.
For a verifier, a certificate image that passes a detector should not be taken as genuine, so detection can support verification with the issuing institution but cannot replace it.

\paragraph{Future work.}
TamperCert can be extended to other text fields such as the degree and the date, to names in non-Latin scripts, to printed and scanned certificates, and to other credentials.
A detector that handles both regional edits and full-image regeneration is still missing, and training on TamperCert does not provide one.
Robustness to post-processing also remains open.
Resizing and recapture remove most of the traces that tampering localizers use, whereas the hit rate of Qwen3-VL on regional edits changes little across versions.
Cues that do not depend on such traces, such as whether a name matches the font and layout used by its university, therefore deserve further study.
Visible flaws are not enough, because general-purpose generative models write legible names that the commercial MLLM rarely flags.
Watermarks embedded by a generator or by the issuing institution are a complementary direction.
They do not mark edits made with image-editing software or with locally run models, and whether they survive a screenshot or a phone photograph can be tested with the versions of TamperCert.

\paragraph{Ethical considerations.}
The genuine certificates come from publicly available sources. All 11 tampering methods are existing software or publicly available models, so TamperCert introduces no new way to forge a certificate. To prevent misuse, the data and code will be provided on request for research use only.

\section{Conclusion}
\label{sec:conclusion}
We presented TamperCert, a text-tampering benchmark of 400 real degree certificates in which 11 tampering methods replace the holder's name and every image is provided in five versions, from a lossless file to a phone photograph.
None of the 13 evaluated detectors is reliable for both regional edits and full-image regeneration, and after resizing alone tampering localizers miss most regional edits.
In phone photographs, text replaced with Photoshop or general-purpose generative models is not reliably detected by any of them.
Training on TamperCert helps for the regional-editing methods seen in training but not for full-image regeneration.
TamperCert makes it possible to evaluate detectors on post-processed images of genuine and forged certificates, and not only on edited files.

% Mandatory for CVPR 2027: after the main text, before the references; does not count
% toward the 8-page limit; must not reveal author identity.
% Mandatory section (CVPR 2027 Author Guidelines): titled "LLM Usage Statement", placed after the
% main text and before the references; not counted toward the page limit; must stay anonymous.
\section*{LLM Usage Statement}
LLMs and other generative AI tools were used in three ways.
First, Claude and GPT were used to polish the wording of the text.
Second, generative models were used to produce data: Qwen-Image-Edit, Gemini and GPT-Image generated forged certificates as tampering methods (\cref{sec:editing}), and Qwen3-VL and Claude were evaluated as detectors (\cref{sec:protocol}).
Third, Claude and GPT assisted in writing the code for the experiments.
The authors take full responsibility for the content of the paper.

{
    \small
    \bibliographystyle{ieeenat_fullname}
    \bibliography{refs}
}

% Supplementary material is a separate file (supplementary.tex), uploaded as a separate
% PDF/ZIP by the supplementary deadline. Do not append it to the main PDF.

\end{document}